\documentclass[num-refs]{wiley-article}
\usepackage{siunitx}
\usepackage{algorithm}
\usepackage{algorithmic}
\usepackage{lmodern}
\usepackage{tabularx} 
\usepackage{color}
\usepackage{diagbox}
\usepackage{caption}
\usepackage{xcolor}
\usepackage{multirow}
\usepackage{subcaption}
\usepackage{adjustbox}
\usepackage{hyperref}
\usepackage{ulem}
\usepackage{soul,color}
\soulregister\cite7
\soulregister\ref7
\soulregister\pageref7
\papertype{Original Article}
\paperfield{Journal Section}
\title{Enhancing Psychologists' Understanding through Explainable Deep Learning Framework for ADHD Diagnosis}

\author[1]{Abdul Rehman}
\author[1]{Jerry Chun-Wei Lin}
\author[1]{Ilona Heldal}

\affil[1]{Department of Computer Science, Electrical Engineering and Mathematical Sciences, Western Norway University of Applied Sciences, Bergen, Norway}

\corraddress{Abdul Rehman and Jerry Chun-wei Lin, Department of Computer Science, Electrical Engineering and Mathematical Sciences, Western Norway University of Applied Sciences, Bergen, Norway}
\corremail{arj@hvl.no, jerry.chun-wei.lin@hvl.no}

\fundinginfo{The research leading to these results is in the frame of the "EMPOWER. Design and evaluation of technological support tools to empower stakeholders in digital education" project, which has received funding from the European Union's Horizon Europe programme under grant agreement No 101060918. Views and opinions expressed are, however, those of the authors(s) only and do not necessarily reflect those of the European Union. Neither the European Union nor the granting authority can be held responsible for them.}

\begin{document}

\maketitle

\begin{abstract}

Attention Deficit Hyperactivity Disorder (ADHD) is a neurodevelopmental disorder that is challenging to diagnose and requires advanced approaches for reliable and transparent identification and classification. It is characterized by a pattern of inattention, hyperactivity and impulsivity that is more severe and more frequent than in individuals with a comparable level of development. In this paper, an explainable framework based on a fine-tuned hybrid Deep Neural Network (DNN) and Recurrent Neural Network (RNN) called \textit{HyExDNN-RNN} model is proposed for ADHD detection, multi-class categorization, and decision interpretation. This framework not only detects ADHD, but also provides interpretable insights into the diagnostic process so that psychologists can better understand and trust the results of the diagnosis.
We use the Pearson correlation coefficient for optimal feature selection and machine and deep learning models for experimental analysis and comparison. We use a standardized technique for feature reduction, model selection and interpretation to accurately determine the diagnosis rate and ensure the interpretability of the proposed framework. Our framework provided excellent results on binary classification, with \textit{HyExDNN-RNN} achieving an F1 score of 99\% and 94.2\% on multi-class categorization. XAI approaches, in particular SHapley Additive exPlanations (SHAP) and Permutation Feature Importance (PFI), provided important insights into the importance of features and the decision logic of models. By combining AI with human expertise, we aim to bridge the gap between advanced computational techniques and practical psychological applications. These results demonstrate the potential of our framework to assist in ADHD diagnosis and interpretation.


\end{abstract}

\section{Introduction} 
ADHD is a neurodevelopmental condition that impacts individuals across the lifespan, encompassing both children and adults~\cite{gu2022adhd}. The condition is characterized by a consistent and enduring pattern of inattention, hyperactivity or impulsivity that can interfere with everyday life and hinder overall development. People with ADHD often show symptoms of inattention (lack of focus), hyperactivity (excessive movement that is inappropriate for sitting) or impulsive behavior (the tendency to act without considering the possible consequences of one's actions). Focusing problems, irritation, distraction and other abnormal mental states are all symptoms of mental disorders~\cite{berrezueta2021robotic}. There is a need to effectively treat mental illness without limiting factors such as lack of knowledge, time or doctor access. Better and more modern patient care can be provided through the popular trend of establishing remote connections between patients and medical experts~\cite{pei2022data}.

Artificial Intelligence (AI) is already playing an important role in several healthcare sectors, including understanding, diagnosing, managing and treating disorders such as ADHD~\cite{rahman2023ai}. AI can analyze medical records, behavioral data and brain imaging techniques (such as functional Magnetic Resonance Imaging (fMRI) scans) to identify patterns associated with ADHD. Indicators of ADHD can be found in small alterations in brain structure or function that machine learning algorithms can efficiently identify. The key to successful intervention and therapy lies in early detection and diagnosis. AI can help find people with ADHD, especially children who may not have severe symptoms but still need help~\cite{sibley2023developing}.

In this paper, an explainable framework based on a fine-tuned hybrid Deep Neural Network (DNN) and Recurrent Neural Network (RNN) named \textit{HyExDNN-RNN} is proposed for ADHD detection and decision interpretation. It utilizes the publicly available ADHD dataset~\cite{bellec2017neuro} potential for binary and multi-class classification problems by combining feature selection and machine and deep learning techniques with XAI methodologies. This framework not only detects ADHD, but also provides interpretable insights into the diagnostic process so that psychologists can better understand the results. In this work, a feature reduction technique is used to select the most relevant features from the dataset to improve the efficiency of the modeling process. Furthermore, in addition to the classification models, we also use XAI techniques such as SHAP~\cite{lundberg2017unified} and PFI~\cite{altmann2010permutation} to interpret the decisions made by the classification model. These XAI techniques shed light on the inner workings of the model and provide insights into the relevance of the features and the decision logic of the model. The proposed framework provided excellent results in binary classification, with the \textit{HyExDNN-RNN} classifier achieving 99\% accuracy and 94.2\% in multi-class categorization. This comprehensive framework improves the interpretability of the models and leads to a better understanding of ADHD identification and classification, expanding knowledge in the field.

This paper is structured as follows: Section \ref{lr} presents related work on the detection and diagnosis of ADHD. Section \ref{pm} presents the proposed framework for explainable ADHD detection and multi-class classification. Section \ref{ev} provides the experimental settings, analyzes and results. Section \ref{Discussion and Findings} discusses the results. Finally, section \ref{conclusion} concludes the paper.

\section{Related Work}\label{lr}
This section presents a literature analysis regarding utilizing XAI techniques in the field of deep learning to identify Attention-Deficit/Hyperactivity Disorder (ADHD).

Deep learning has shown exceptional capabilities in processing and analyzing complex data, especially when using neural networks such as Convolutional Neural Networks (CNNs)~\cite{o2015introduction} and RNNs~\cite{wang2022attention}. Zhang et al.~\cite{zhang2020separated} laid the foundation for the application of machine learning techniques in the classification of ADHD. They used structural neuroimaging data in their study and demonstrated the suitability of deep learning (CNN) techniques to differentiate individuals with ADHD. CNN is then used to learn the temporal features of individual brain areas. In the second stage, a fusion feature extraction network detects the temporal relationships between these areas. The two-stage design effectively utilizes the inherent temporal features and interactions of whole-brain resting-state fMRI data. The method uses a "leave-one-site-out" cross-validation procedure to achieve an average classification accuracy of 68.6\% for the five sites. The results show that the proposed network is stable and reliable across multiple sites, although the data varies. Finally, it is shown that combining CNN with the attention network is an effective method to utilize intrinsic fMRI information to discriminate ADHD in resting-state fMRI datasets from multiple sites~\cite{zhang2020separated}. 

Understanding the decision-making processes in deep learning models is crucial, especially in a therapeutic setting, as they tend to act as transparent entities. XAI approaches, such as SHAP and PFI, have been developed to explore the inner workings of these transparent systems. Lundberg et al.~\cite{lundberg2017unified} have established SHAP values as a comprehensive framework for mapping features in intricate models. Identifying the most informative features from large datasets can significantly improve the accuracy of categorization. Mao et al ~\cite{mao2019spatio} emphasizes the importance of feature reduction strategies. The use of feature selection with deep learning was investigated in ADHD classification to improve the model efficiency of the model they developed. The integration of deep learning models with XAI provides a comprehensive framework for developing interpretable AI in healthcare. Huang et al. demonstrated the effective integration of SHAP values into deep learning algorithms for medical analysis. This integration makes it easier for clinicians to develop confidence in and understand the predictions~\cite{huang2019diagnosis}.

%% I dont think we need this in the literature review
%To obtain better explainability, we improve model transparency and interpretability through feature reduction and XAI techniques (i.e., SHAP and PFI). Incorporating XAI methods and deep learning models, our unique strategy for ADHD classification showed explainable performance in ADHD detection. The performance experiments of the developed framework are promising in binary classification tasks.

\section{Proposed Framework} \label{pm}
\begin{figure*}[!ht]
\centering
\includegraphics[width=\textwidth]{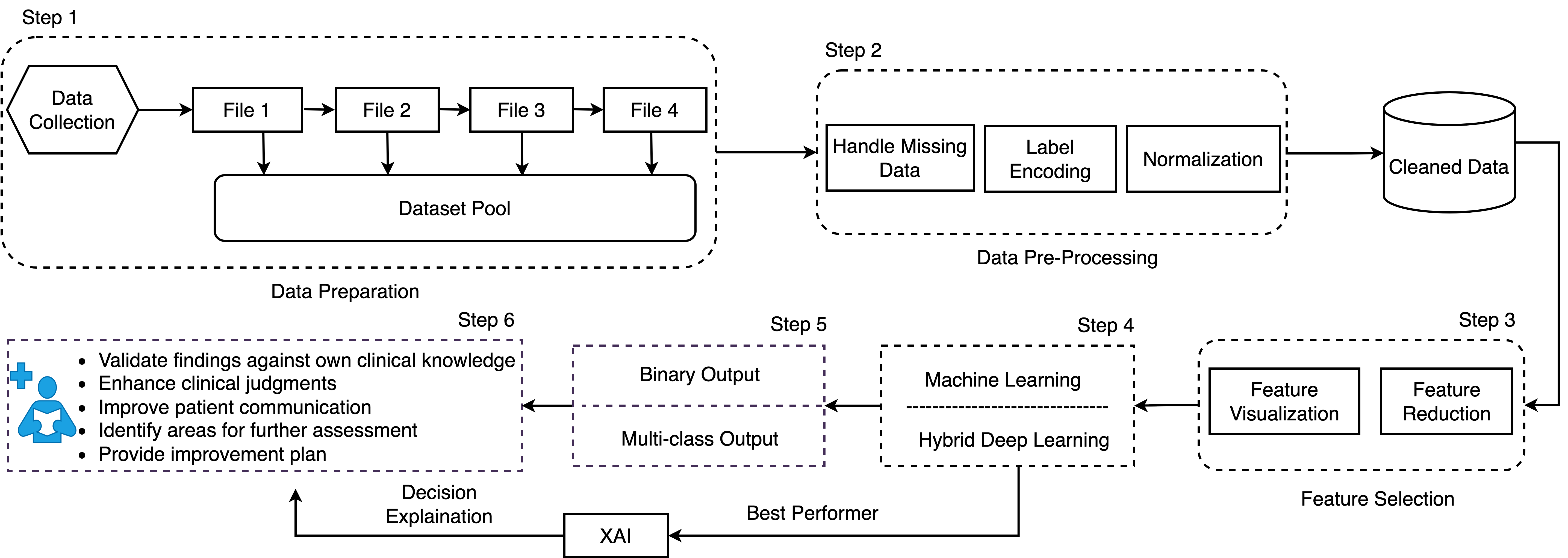}
\caption{Proposed XAI-based Machine and Deep Learning Framework for ADHD Diagnosis and Explainability}
\label{proposed_digram}
\end{figure*}

The proposed \textit{HyExDNN-RNN} framework aims to improve the efficiency and interpretability of ADHD detection by integrating XAI techniques with deep learning models. Furthermore, it aims to elucidate the efficiency of XAI-driven deep learning models in addressing the challenges associated with ADHD diagnosis. This framework not only detects ADHD but also provides interpretable insights into the diagnostic process so that psychologists can better understand the results. The steps of the proposed methodology are illustrated in Figure \ref{proposed_digram}. It starts with the selection of the dataset, data preparation and preprocessing. Next, the feature reduction method selectively extracts the most important features from the dataset to streamline the modeling process and increase efficiency. Next, various machine and deep learning models (i.e. LSTM, Deep Neural Network (DNN), LSTM-Gated Recurrent Unit (LSTM-GRU), DNN-RNN, LSTM-RNN, Random Forest (RF), Decision Tree (DT) and Extreme Gradient Boosting (EXGB)) are used to perform classification tasks. The XAI tools are then used to demonstrate the performance of the proposed model and provide important insights into the meaning of the features and the reasons for the model's decisions.

Algorithm \ref{algo:adhd-diagnosis} presents the steps of the proposed framework for ADHD diagnosis, which includes data preprocessing, feature reduction, the use of different machine learning models, a deep learning model (LSTM-RNN), and a comprehensive explainability analysis to provide insights into the models' decisions and support the diagnosis process. Data preparation and preprocessing focus on preparing the input data for the subsequent stages. This includes dealing with missing data, a crucial step in data pre-processing. This includes techniques such as imputation or the removal of data points with missing values. It also performs label encoding, which converts categorical labels into numerical values, and normalization of the data using a standard scaler to ensure that the features have a similar scale, which can be important for some machine learning algorithms. Next, the algorithm selects the most important features from the dataset using the Pearson correlation coefficient ~\cite{cohen2009pearson}. This helps to reduce the dimensionality of the data while retaining the most important information. After feature reduction, the algorithm trains machine learning models (Random Forest - RF, Decision Tree - DT and Extreme Gradient Boosting - EXGB) on the reduced feature set. In addition to the traditional machine learning models, the algorithm trains a \textit{HyExDNN-RNN} model on the entire dataset and other deep learning models such as DNN, LSTM, LSTM-GRU and LSTM-RNN. Furthermore, the algorithm performs an explainability analysis for the \textit{HyExDNN-RNN}, calculates the importance of local features based on SHAP values and visualizes the local explanation. Local explanations help to understand why a certain prediction was made for an individual.
In addition, the importance of global features is calculated using PFI to identify the most important features for all predictions of the developed model. This is often useful to understand the overall behavior of the model. The explainable features step identifies important features that contribute to the diagnosis and illustrates how these features influence the model's decision. Finally, the algorithm predicts the ADHD diagnosis using all trained models for each subject. The results show the findings of the diagnosis, which can be considered as information for a psychologist to better understand the results and their explainability.

\begin{algorithm}[!ht]
\caption{Pseudo code for XAI-based ADHD Diagnosis Framework}
\label{algo:adhd-diagnosis}
\begin{algorithmic}[1]
\STATE \textbf{Input: $\text{ADHD200\_Data}$}
\STATE \textbf{Output: ADHD Diagnosis, Explanation}
\STATE \textbf{Data Pre-processing}
\STATE \textbf{Feature Reduction:} Select the most important features using the Pearson correlation coefficient\;
\STATE \textbf{Machine and Deep Learning Models:} 
Train machine learning models (i.e., RF, DT, EXGB) and DNN, LSTM, LSTM-GRU, LSTM-RNN, \textit{HyExDNN-RNN} model on the reduced feature set\;
\STATE \textbf{Explainability Analysis():} 
 \FOR{Model}
 \STATE \textbf{Local Explanations:} Select a subject data point $X_i$\; 
\STATE Compute local feature importance for $X_i$ using SHAP values\;
\STATE \textbf{Global Explanations:} Calculate global feature importance using PFI\;
\STATE \textbf{Model Visualization:} Visualize layers and neurons\;
 \STATE \textbf{Explainable Features:} Identify important features contributing to the diagnosis\;
\STATE Visualize the impact of features on the model's decision\;
\ENDFOR
\FOR{subject data point $X_i$}
 \STATE Predict ADHD diagnosis using all models\;
 \IF{Diagnosis is Positive} 
 \STATE Show the explainability results\;
 \ENDIF
\ENDFOR
\RETURN Psychologists $\gets$ Diagnosis, Explanations
\STATE Psychologists provides:
\STATE \textbullet{} Validate Findings against Own Clinical Knowledge
\STATE  \textbullet{} Enhance Clinical Judgments
\STATE  \textbullet{} Improve Patient Communication
\STATE  \textbullet{} Identify Areas for Further Assessment
\STATE  \textbullet{} Provide Improvement Plan
\end{algorithmic}
\end{algorithm}

\subsection{Dataset Selection}
\begin{figure}[!ht]
\centering
\includegraphics[width=0.6\columnwidth]{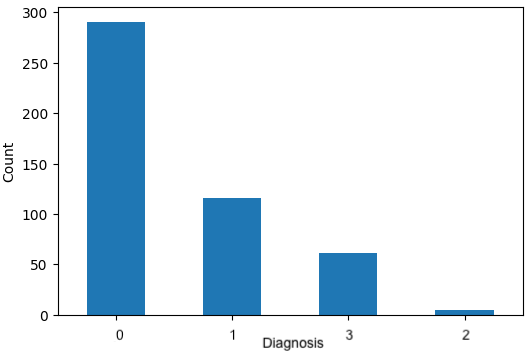}
\caption{Distribution of Dataset Classes. Typically Developing Children (TDC) Class 0, ADHD-combined (Class 1), ADHD-Hyperactive/Impulsive (Class 2), ADHD-Inattentive (Class 3)}
\label{final}
\end{figure}

This paper uses the ADHD200 dataset~\cite{bellec2017neuro}, which includes numerous attributes related to the subjects and their ADHD diagnosis. These attributes include ScanDir ID, Site, Gender, Age, Handedness, Diagnosis (DX), Secondary Diagnosis (Dx), ADHD Measure, ADHD Index, Inattentive, Hyper/Impulsive, IQ Measure, Verbal IQ, Performance IQ, Full2 IQ, Full4 IQ, Med Status, and several Quality Control (QC) measures for resting state and anatomical data. The data collection includes information from many sites and represents people with different demographic and clinical features. DX distinguishes between people with ADHD (1) and healthy controls (0), while Secondary Dx provides additional diagnostic information. ADHD measures such as inattentiveness and hyperactivity/impulsivity are associated with ADHD symptoms.
In addition, IQ and Med Status assessments provide information about participants' cognitive abilities and medication status. The QC metrics indicate the quality of resting state and anatomical data. Figure \ref{final} shows the distribution of classes in both the binary and multi-class scenarios. The distribution of classes has Developing Children (TDC) (Class 0), which is the most common in this dataset, with 291 instances, followed by ADHD-combined (Class 1) with 116. ADHD - Inattentive (Class 3) represents 61 instances, while ADHD - Hyperactive/Impulsive (Class 2) has only 5 examples. This class distribution within the dataset provides useful information about the composition of the dataset and is crucial for identifying potential class imbalances that may affect classification.

\subsection{Data Pre-Processing}
The steps for data preparation and preprocessing are described in detail in this section. The dataset is first divided into four subgroups, and the first goal is to integrate these subsets into a single file for unified analysis. However, it is important to note that this combined dataset contains some values that need to be imputed. To solve the problem of missing data, a common strategy was chosen. The missing values are replaced by the mean value of the relevant columns. This imputation approach is suitable if it is assumed that missing data occurs randomly and that filling with the mean value is an acceptable approximation. Label encoding is used to make the dataset compatible with machine learning methods. This transformation converts the values in the columns of the object data type into unique integer values. Many machine learning models require numeric inputs, so this is crucial. Finally, the data is normalized using the default scaler to ensure consistency and that all features are on the same scale. Normalization prevents some traits from dominating the modeling process due to scale disparities.

\subsection{Feature Selection}
The Pearson correlation coefficient~\cite{cohen2009pearson} is a beneficial statistical tool to examine the correlations between different variables in the dataset. The Pearson correlation measures the degree and direction of linear relationships between two variables. It assesses how strongly changes in one variable are reflected in changes in another, with values between [0, 1]. The main objective of using Pearson correlation is to examine the degree of relationship between the features and to select the features with unique and non-redundant information. If features have a high correlation, this indicates that they may contain comparable information. Adding both features for analysis may lead to multicollinearity, which may affect the performance and interpretability of the model. To address these concerns, a 95\% criterion is set for feature removal. Features with a correlation above this value are considered highly correlated and are, therefore, redundant for further experimentation.

\subsection{Research Preliminary}
In this section, the mathematical models underlying the ADHD frameworks are presented. The proposed framework uses the XAI method to generate acceptable and intuitive explanations that can be used to examine the functioning of the model, the quality of the datasets, and the features used in the numerous ADHD datasets. LSTM is a form of the well-known recurrent neural network (RNN), which is known for its ability to process sequential data. We use a fine-tuned DNN-RNN model called \textit{HyExDNN-RNN} to capture subtle patterns and dependencies in sequence data, making it suitable for the dynamic nature of ADHD. The designed \textit{HyExDNN-RNN} model consists of multiple sequential stacks of layers and starts with two dense layers with 256 units and 128 neurons with ReLU activation, respectively, which process the input sequence while maintaining structure. After the dense layers, the model includes a Dropout layer to reduce overfitting by randomly setting a fraction (0.5) of the input units to 0 during the training phase. Next, two RNN layers with 64 and 32 units, respectively, are added. The first RNN layer has return\_sequences=True, which outputs the complete sequence for the next layer. After the first RNN layer, another Dropout layer is added for regularization. Finally, the model contains a dense layer with 16 neurons and ReLU activation, followed by an output layer with 4 neurons using the softmax activation function, which is suitable for multi-class classification. The model is compiled using the Adam optimizer and categorical cross-entropy loss, a standard choice for multi-class classification problems. The model is compiled using the binary cross-entropy loss function and the Adam optimizer and is set to track accuracy as a performance measure. The model runs through 120 epochs during the training phase, processing the data in batches of 32 samples. For binary labelling, the dataset is split into 75\% for training and the rest for testing. For multi-class labelling, the dataset is split into 80\% for training and the remainder for testing. A validation split of 20\% is also used to evaluate the performance of the model and avoid overfitting. One of the most important components of this XAI-based LSTM model is the use of PFI~\cite{altmann2010permutation}. This technique evaluates the impact of randomly permuting the values of each feature on the performance of the model to determine the relevance of certain features. As a result, each feature is given a list of relevant values that help to clarify the relative importance of different input factors in the decision-making process of the developed model.

\section{Experimental Evaluation}
\label{ev}
This section contains the results and experimental setup used for the ADHD dataset. This work aims to analyze ADHD using machine learning and deep learning models and divide it into different categories, both binary (ADHD or Normal). The proposed framework focuses on the detection of ADHD markers within a dataset while minimizing the effects of noise and confounding factors. Three independent machine learning models, RF~\cite{breiman2001random}, DT \footnote{https://scikit-learn.org/stable/modules/tree.html} and EXGB\footnote{https://github.com/dmlc/xgboost}, are used for this purpose, along with deep learning models: DNN, LSTM, LSTM-RNN, LSTM-GRU. In addition, XAI approaches such as SHAP and PFI are used to gain deeper insights into the decision process of the model and to show the importance of the developed model. SHAP values show how each feature contributes to the model's predictions. They allow the identification of key factors that have a significant impact on the classification of ADHD groups. By visualizing the SHAP scores, it is possible to understand and explain the results of the model and shed light on the aspects that have the greatest impact on the results. PFI, on the other hand, helps to determine the sensitivity of the model to changes in certain features. It determines which features affect the accuracy of the model the most when they are perturbed or swapped. This analysis helps to identify the robustness of the model and provides information about the relevance of the features. The use of XAI approaches in conjunction with the deep learning LSTM model improves the interpretability and transparency of the ADHD classification process. This comprehensive technique produces accurate results and provides insight into the critical traits and decision logic underlying the model's classifications. It advances the application of deep learning in medical classification tasks and contributes to a deeper understanding of ADHD diagnosis.

\subsection{Experimental Settings}
This section describes the experimental setting used to conduct the investigation. The experiments are conducted on the cloud-based Kaggle platform \footnote{https://www.kaggle.com/}, which enables access to GPU resources for greater computing capabilities. Python version 3.8.8 ensures compatibility with the chosen libraries and tools for data analysis and machine learning. We evaluate the proposed approach using various metrics such as accuracy ($\text{Accuracy} = \frac{\text{Number of Correct Predictions}}{\text{Total Number of Predictions}}$), precision ($\text{Precision} = \frac{\text{TP}}{\text{TP + FP}}$), recall ($\text{Recall} = \frac{\text{TP}}{\text{TP + FN}}$), and F1-score ($\text{F1-Score} = 2 \times \frac{\text{Precision} \times \text{Recall}}{\text{Precision + Recall}}$). The above formulas use True Positives (TP), True Negatives (TN), False Positives (FP), and False Negatives (FN) to calculate these metrics.

\subsection{Experimental Results}

\subsubsection{Binary Classification Results}

Table \ref{table1} presents the results of the different binary classification models in terms of key performance metrics: Accuracy, Precision, Recall, F1-score, Receiver Operating Characteristic- Area Under Curve (ROC-AUC). Each model is evaluated to determine its effectiveness in correctly classifying data into two categories. RF achieved an accuracy, precision, recall, and F1-score of 96\%. This indicates a high level of performance as positive and negative instances are correctly identified with consistent reliability. DT showed an accuracy, precision, recall and F1 score of 96\%, matching the performance of the RF model. This suggests that DT is equally effective in this classification task and offers a similar level of correctness and reliability. EXGB stands out with slightly better performance, achieving 97\% on all metrics. This model shows a superior ability to accurately classify the instances, making it one of the top performers in this evaluation. The LSTMs achieved an accuracy, precision, recall, and F1-score of 90\%. Although it still performs quite well, it falls behind the other models, suggesting that it may not be as effective for this particular binary classification task. DNN achieved 90\% on all metrics. It is slightly less accurate than EXGB and LSTM-RNN, LSTM-GRU, and \textit{HyExDNN-RNN}, but still shows high reliability and effectiveness. LSTM-GRU also performed well, with 95\% for all metrics.
LSTM-RNN achieved the same performance as EXGB, namely 97\% for accuracy, precision, recall, and F1-score. When comparing the models, \textit{HyExDNN-RNN} emerged as the frontrunner, achieving 99\% in all metrics. The LSTM-GRU model also performed well with 95\%, making it a strong candidate. The LSTM model lagged behind the others with 90\%, suggesting that it may not be as well suited for this particular binary classification task. Overall, based on the results, \textit{HyExDNN-RNN} is recommended for its superior accuracy and consistency in performance. These metrics highlight the high performance of the \textit{HyExDNN-RNN} classifier with consistently high precision, recall and F1 scores for both classes and an impressive overall accuracy of 99\%. It should also be noted that the ROC-AUC of \textit{HyExDNN-RNN} is the highest at 0.99.

\begin{table}[!ht]
\centering
\caption{Results for Binary Classification (Weighted Average)}
\label{table1}
\begin{tabular}{|c|c|c|c|c|c|}
\hline
\textbf{Models} & \textbf{Accuracy (\%)} & \textbf{Precision (\%)} & \textbf{Recall (\%)} & \textbf{F1-score (\%)} & \textbf{ROC-AUC}\\ \hline
RF &96.0 &96.0&96.0&96.0& 0.97\\ \hline
DT &96.0&96.0&96.0&96.0&0.95 \\ \hline
EXGB &97.0&97.0&97.0&97.0&0.97 \\ \hline
LSTM & 90.0&90.0&90.0&90.0&0.92 \\ \hline
DNN &90.0 &90.0&90.0&90.0& 0.95\\ \hline
LSTM-GRU &95.0&95.0&95.0&95.0&0.90 \\ \hline
LSTM-RNN &97.0 &97.0&97.0&97.0&0.94 \\ \hline
\textit{HyExDNN-RNN} & 99.0 & 99.0 & 99.0 & 99.0&0.99 \\ \hline
\end{tabular}
\end{table}

\begin{figure*}[!ht]
\centering
\subfloat[Training and Validation Accuracy of LSTM\label{figlstm(c)}]{
\includegraphics[width=0.5\textwidth]{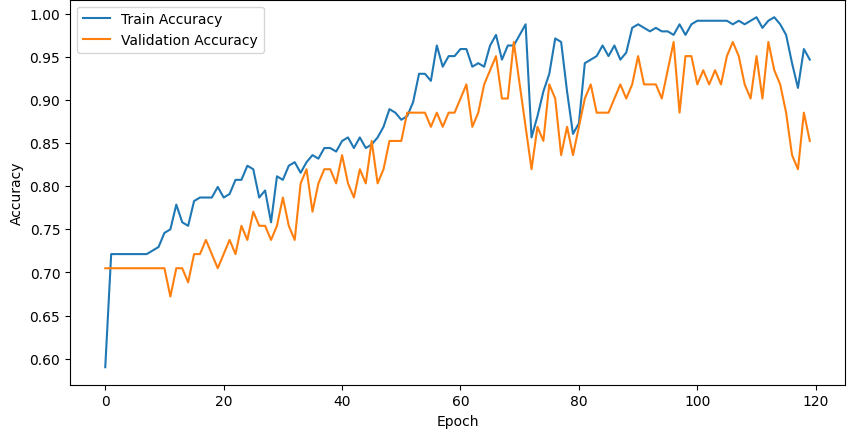}}
\subfloat[Training and Validation Loss of LSTM\label{figlstm(d)}]{
\includegraphics[width=0.5\textwidth]{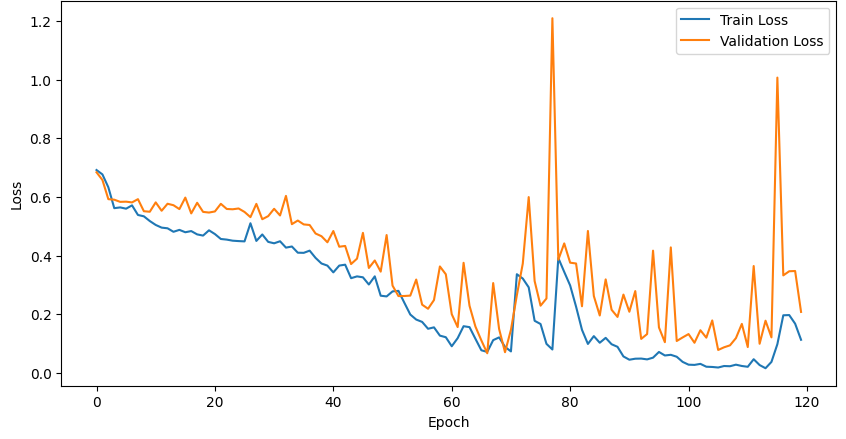}}
\caption{Graphical Visualization of LSTM Model Results for Binary Classification}
\label{figlstm}
\end{figure*}

The performance of the LSTM model is shown graphically in Figure \ref{figlstm}. The training and validation accuracy of an LSTM model over 120 epochs is shown in Figure \ref{figlstm}(a). The number of epochs used to train the model is shown on the x-axis. The accuracy of the correct predictions is shown on the y-axis. Both training and validation accuracy are low in the early stages of the training process. The training accuracy value starts at 0.60 in the first epoch and reaches a value of 0.98 in the $119_{th}$ epoch after some fluctuations of increase and decrease. The value of the training accuracy decreases to 0.95 at $120_{th}$ epoch. The validation accuracy starts at 0.70 in the first epoch and reaches a value of 0.95 in the $75_{th}$ epoch after some fluctuations of increase and decrease. The value of the validation accuracy decreases and ends at 0.85 in the $120_{th}$ epoch. The training and validation loss of an LSTM model over ten epochs is shown in Figure \ref{figlstm}(b). The training loss starts at 0.7 in the first epoch and ends after some fluctuations at $120_{th}$ epoch with a value of 0.15. The validation loss starts at 0.7 in the first epoch and reaches a value of 1.19 at $79_{th}$ epoch. After some fluctuations of increase and decrease, it then stops at 0.2 at $120_{th}$ epoch.

\begin{figure*}[!ht]
\centering
\subfloat[Training and Validation Accuracy of DNN\label{fig2(c)}]{
\includegraphics[width=0.5\textwidth]{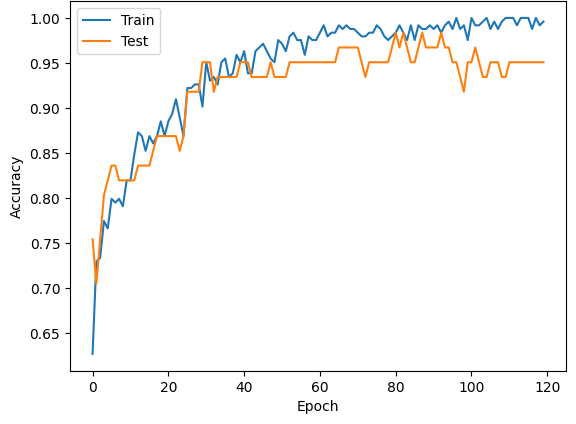}}
\subfloat[Training and Validation Loss of DNN\label{fig2(d)}]{
\includegraphics[width=0.5\textwidth]{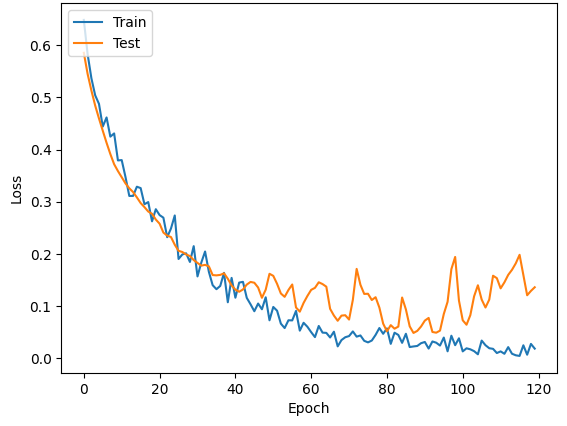}}
\caption{Graphical Visualization of DNN Model Results for Binary Classification}
\label{fig2}
\end{figure*}

The performance of the DNN model is shown graphically in Figure \ref{fig2}. The training and validation accuracy of a DNN model over 120 epochs is shown in Figure \ref{fig2}(a). The value for the training accuracy starts at 0.60 in the first epoch and reaches a value of 0.98 in the $120_{th}$ epoch after some fluctuations of increase and decrease. The validation accuracy starts at 0.75 in the first epoch and reaches a value of 0.95 in the $116_{th}$ epoch after some fluctuations in the increase and decrease. The value of the validation accuracy remains constant until $120_{th}$ epoch. The training and validation loss of a DNN model over 120 epochs is shown in Figure \ref{fig2}(b). The training loss starts at 0.7 in the first epoch and ends after some fluctuations at $120_{th}$ epoch with a value of 0.03. The validation loss starts at 0.59 in the first epoch and ends after fluctuations in increase and decrease at 0.12 in the $120_{th}$ epoch.

\begin{figure*}[!ht]
\centering
\subfloat[Training Accuracy and Validation of LSTM-GRU\label{fig3(c)}]{
\includegraphics[width=0.5\textwidth]{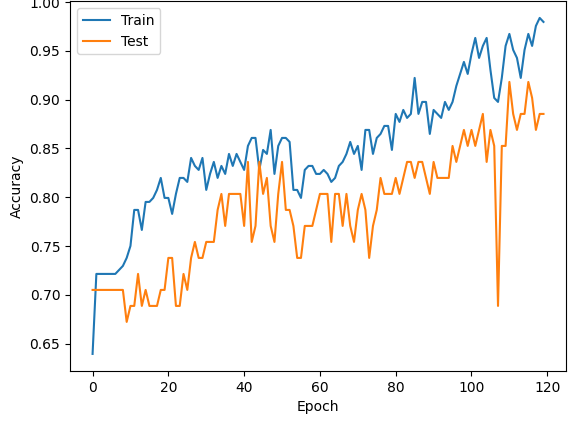}}
\subfloat[Training and Validation Loss of LSTM-GRU\label{fig3(d)}]{
\includegraphics[width=0.5\textwidth]{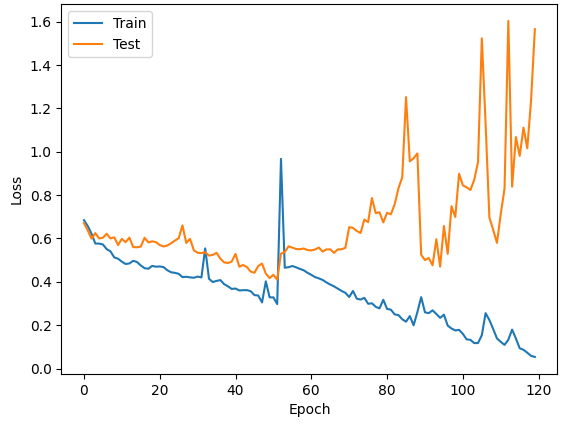}}
\caption{Graphical Visualization of LSTM-GRU Model Results for Binary Classification}
\label{fig3}
\end{figure*}

The performance of the LSTM-GRU model is shown graphically in Figure \ref{fig3}. The training and validation accuracy of an LSTM-GRU model over 120 epochs is shown in Figure \ref{fig3}(a). The value of the training accuracy starts at 0.65 in the first epoch and increases to 0.72 in the same epoch. After some fluctuations between increase and decrease, it reaches a value of 0.98 at the $120_{th}$ epoch. The validation accuracy starts at 0.70 in the first epoch and remains constant until the $8_{th}$ epoch. Then, after some fluctuations of increase and decrease, it drops to 0.70 at the $114_{th}$ epoch. The validation accuracy value starts to increase again, and after some fluctuations, it stops at 0.87 at the $120_{th}$ epoch. The training and validation loss of an LSTM-GRU model over 120 epochs is shown in Figure \ref{fig3}(b). The training loss value starts at 0.7 in the first epoch and increases after some fluctuations with a value of 0.99 in the $50_{th}$ epoch. Then, after some fluctuations, it decreases to 0.05 at $120_{th}$ epoch. The validation loss starts at 0.70 in the first epoch, and after fluctuations in increase and decrease, it ends at 0.15 in the $120_{th}$ epoch.

\begin{figure*}[!ht]
\centering
\subfloat[Training Accuracy and Validation\label{fig4(c)}]{
\includegraphics[width=0.49\textwidth,height=5cm]{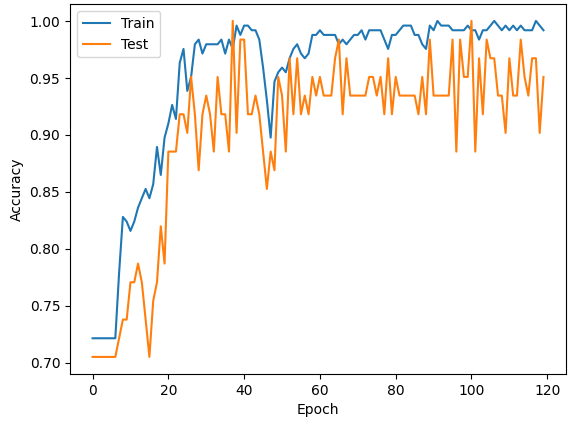}}
\subfloat[Training Accuracy and Validation\label{fig4(d)}]{
\includegraphics[width=0.49\textwidth,height=5cm]{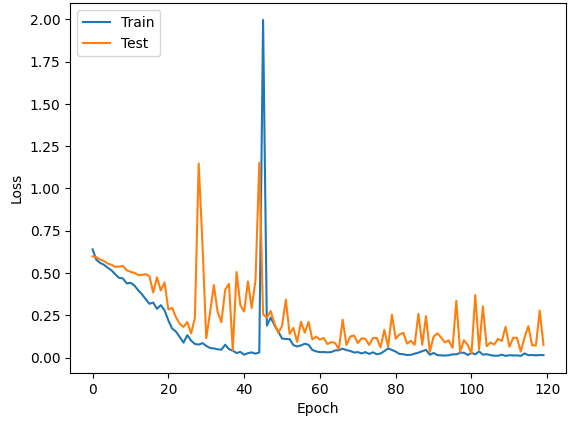}}
\caption{Graphical Visualization of LSTM-RNN Model Results for Binary Classification}
\label{fig4}
\end{figure*}

The performance of the LSTM-RNN model is shown graphically in Figure \ref{fig4}. The training and validation accuracy of an LSTM-RNN model over 120 epochs is shown in Figure \ref{fig4}(a). The training accuracy value starts at 0.72 in the first epoch and remains constant until the $5_{th}$ epoch. After some fluctuations between increase and decrease, it reaches a value of 0.98 at $120_{th}$ epoch. The validation accuracy starts at 0.72 at the first epoch and remains constant until the $6_{th}$ epoch. Then, after some fluctuations of increase and decrease, it stops at 0.94 at the $120_{th}$ epoch. The training and validation loss of an LSTM-RNN model over 120 epochs is shown in Figure \ref{fig4}(b). The training loss value starts at 0.66 in the first epoch and increases after some fluctuations with a value of 2.00 in the $43_{th}$ epoch. Then, after some fluctuations, it decreases to 0.02 at $120_{th}$ epoch. The validation loss starts at 0.60 in the first epoch, and after fluctuations in increase and decrease, it ends at 0.04 in the $120_{th}$ epoch.

\begin{figure*}[!ht]
\centering
\subfloat[Training Accuracy and Validation\label{fig7(c)}]{
\includegraphics[trim={0 0 0 0.7cm},clip,width=0.49\textwidth,height=5cm]{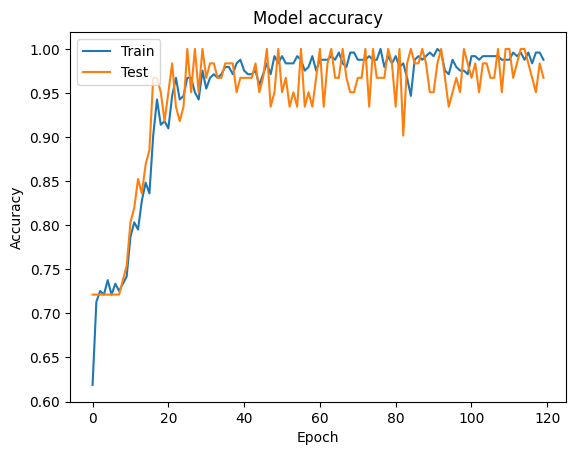}}
\subfloat[Training Accuracy and Validation\label{fig7(d)}]{
\includegraphics[trim={0 0 0 0.7cm},clip,width=0.49\textwidth,height=5cm]{finalimages/DNN-RNN-Accuracy-curve.png}}\\
\subfloat[ADHD features importance extracted by SHAP \label{b_shap}]{
\includegraphics[width=0.49\textwidth,height=7cm]{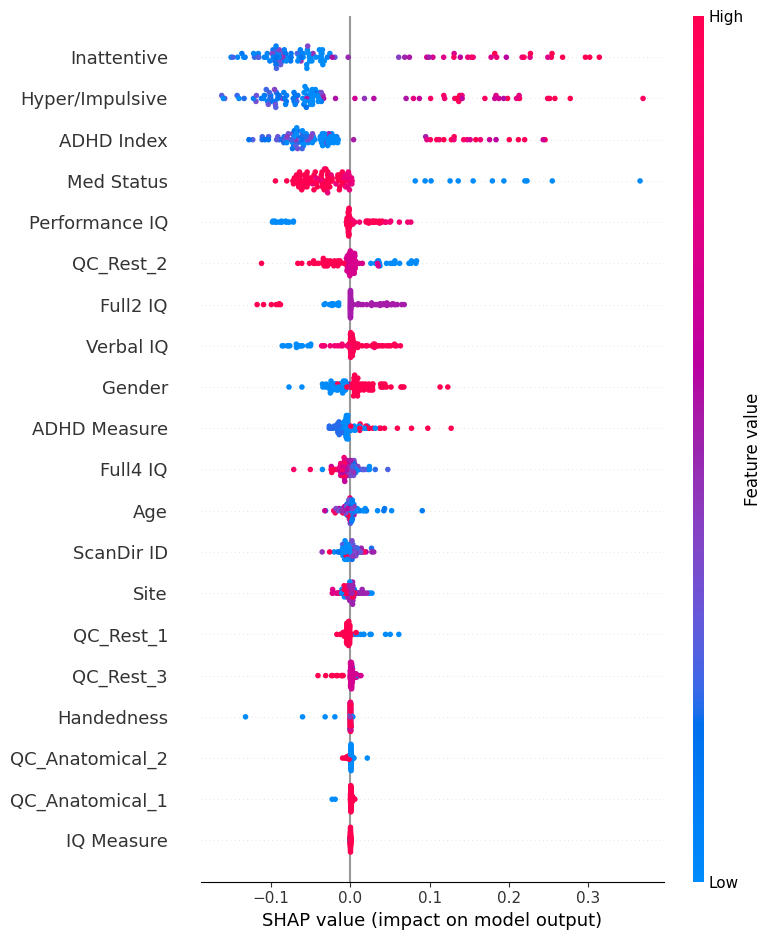}}
\subfloat[ADHD features extracted by PFI \label{b_pfi}]{
\includegraphics[trim={0 0 0 0.7cm},clip,width=0.49\textwidth,height=6cm]{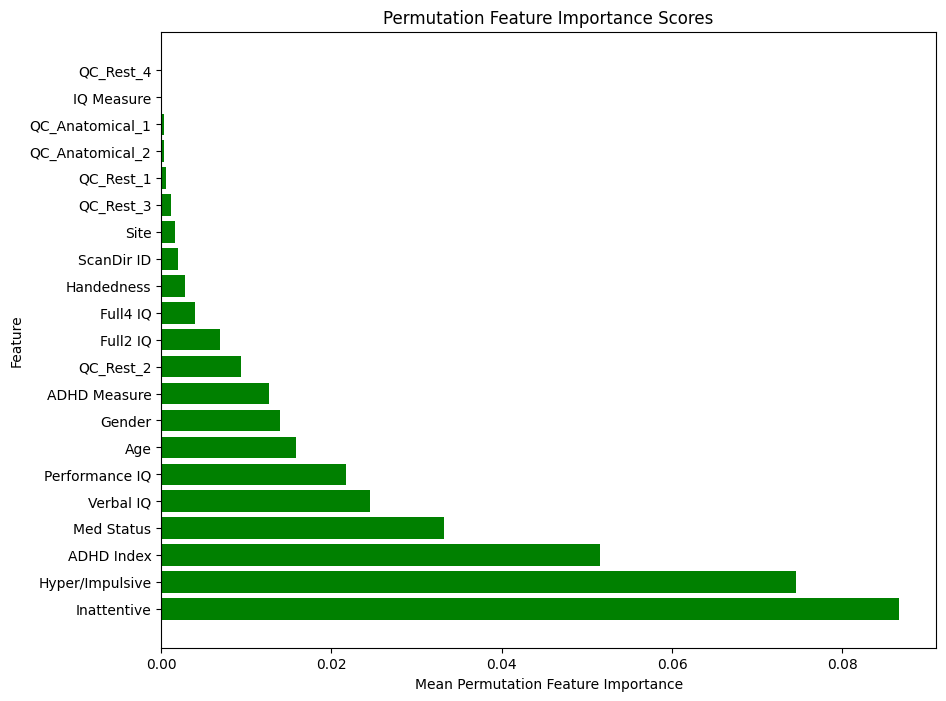}}
\caption{Graphical Visualization of \textit{HyExDNN-RNN} Model Results for Binary Classification}
\label{fig7}
\end{figure*}

The \textit{HyExDNN-RNN} model performance is shown graphically in Figure \ref{fig7}. The training and validation accuracy of a \textit{HyExDNN-RNN} model over 120 epochs is shown in Figure \ref{fig7}(a). The training accuracy value starts at 0.72 in the first epoch and remains constant until the $5_{th}$ epoch. Then, after some fluctuations of increase and decrease, it reaches up to 0.98 at $120_{th}$ epoch. The validation accuracy starts at 0.72 in the first epoch and remains constant until the $6_{th}$ epoch. Then, after some fluctuations of increase and decrease, it stops at 0.94 at the $120_{th}$ epoch. The training and validation loss of a \textit{HyExDNN-RNN} model over 120 epochs is shown in Figure \ref{fig7}(b). The training loss starts at 0.66 in the first epoch and increases after some fluctuations with a value of 2.00 in the $43_{th}$ epoch. Then, after some fluctuations, it decreases to 0.02 at $120_{th}$ epoch. The validation loss starts at 0.60 in the first epoch and ends at 0.04 in the $120_{th}$ epoch after fluctuations.
The SHAP method provides a unique approach for displaying feature importance, as shown in Figure \ref{fig7}(c). This shows the feature importance of feature effects. The vertical position of a point stands for the feature it represents, while its horizontal position indicates the influence of the value on the prediction of the model. High Shapley numbers of the inattentive, hyper/impulsive ADHD index lead to ADHD prediction and are the most important features. In Figure \ref{fig7}(d), the predictions of the XAI-based \textit{HyExDNN-RNN} model provide some interesting facts. It shows the importance of SHAP features measured as mean absolute Shapley values. For example, it also predicts that inattentive, hyper/impulsive and the ADHD index are very important features to predict ADHD. While these features have the greatest impact on the model's predictions, other factors in the dataset also play a role, albeit to a lesser degree. For example, the 'Site' feature, which indicates the location of the scan, and the 'QC\_Rest\_4' feature, which measures a subject's attention and concentration, have an obvious but small impact on the model's predictions. It is important to note that these results indicate the model's projected feature relevance. The features that are most relevant to the model's predictions may differ depending on the features of the dataset and the hyperparameters used in model training. The SHAP provides a more comprehensive understanding of the importance of features and thus addresses the fundamental disadvantage of the PFI explanations mentioned above.

\begin{figure*}[!ht]
\centering
\subfloat[DT model \label{fig8(adasccasc)}]{
\includegraphics[width=0.33\columnwidth,height=4cm]{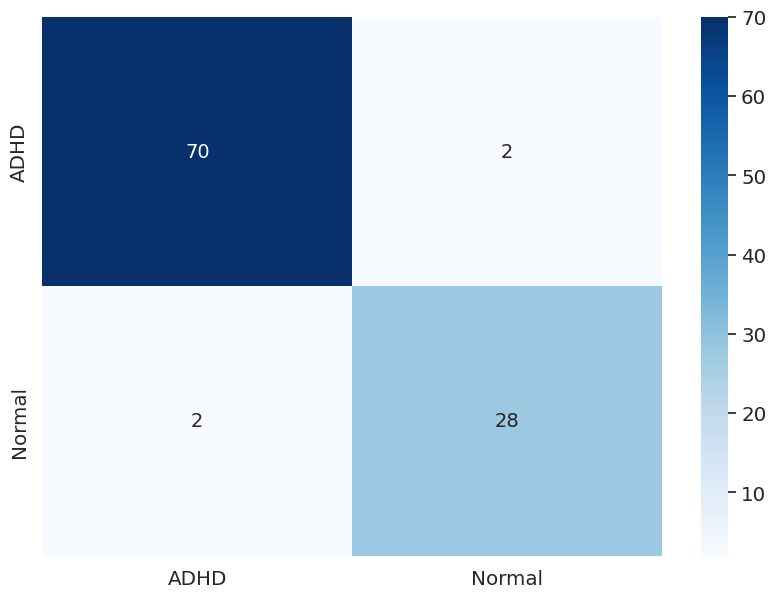}}
\subfloat[RF model \label{fig8(asacdqwdqw)}]{
\includegraphics[width=0.33\columnwidth,height=4cm]{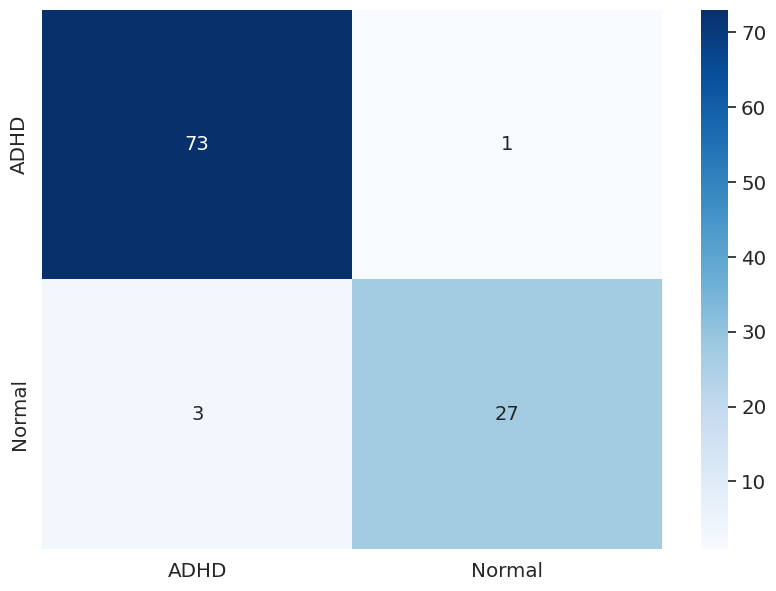}}
\subfloat[EXGB model \label{fig8(aasdada)}]{
\includegraphics[width=0.33\columnwidth,height=4cm]{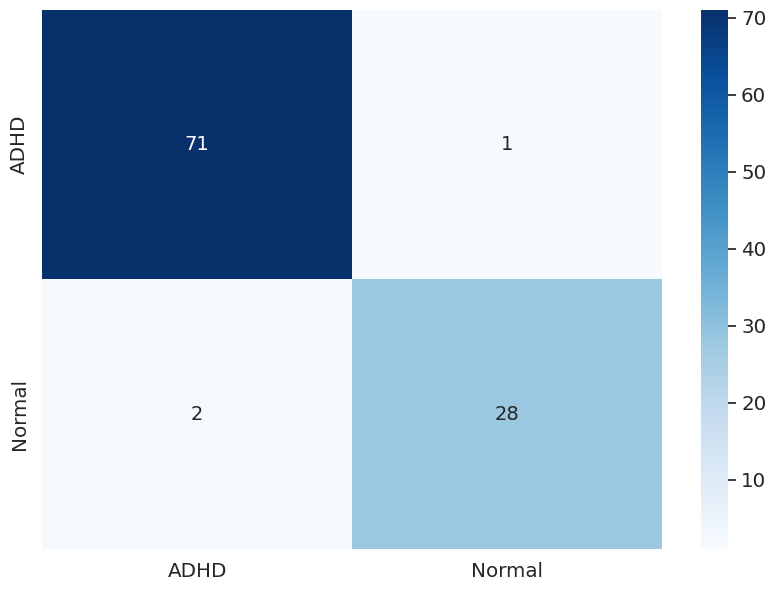}}\\
\subfloat[LSTM model \label{fig8(asac)}]{
\includegraphics[width=0.33\columnwidth,height=4cm]{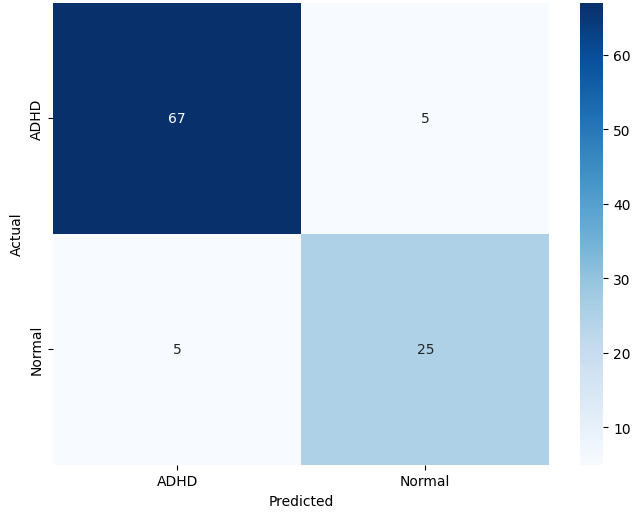}}
\subfloat[DNN model \label{fig8as(azvdsv)}]{
\includegraphics[width=0.33\columnwidth,height=4cm]{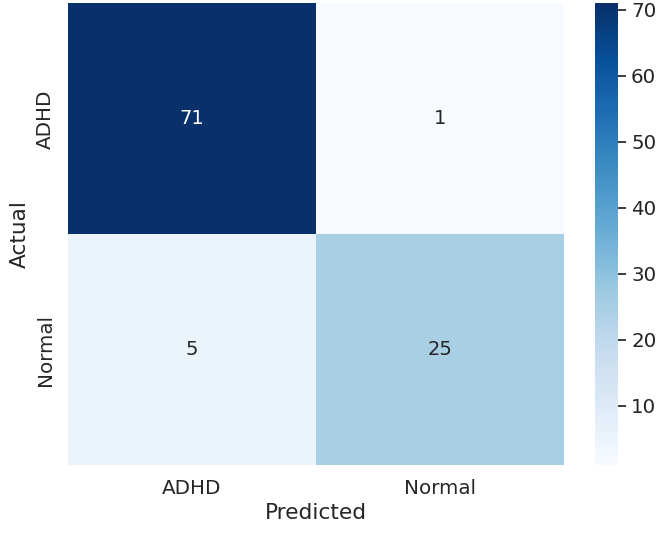}}
\subfloat[LSTM-GRU model \label{fig3(qweqa)}]{
\includegraphics[width=0.33\columnwidth,height=4cm]{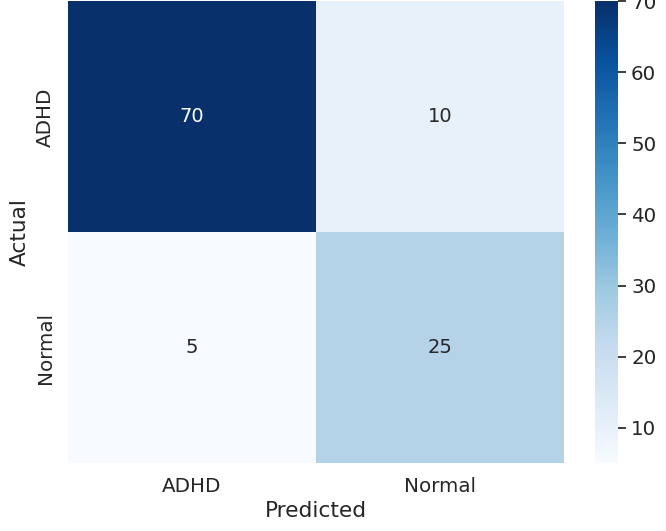}}\\
\subfloat[LSTM-RNN model \label{fig4(qwqwewa)}]{
\includegraphics[width=0.33\columnwidth,height=4cm]{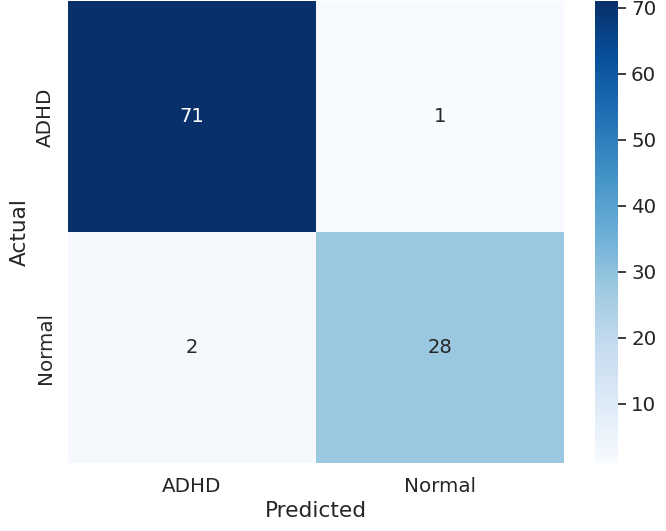}}
\subfloat[\textit{HyExDNN-RNN} model \label{fig4qweqw(a)}]{
\includegraphics[width=0.33\columnwidth,height=4cm]{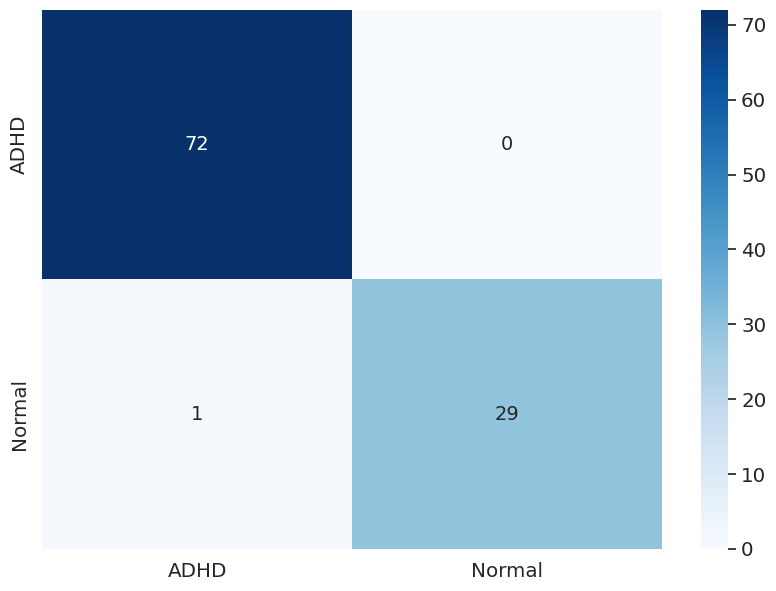}}
\caption{Graphical Visualization of ML Models Results for Binary Classification}
\label{fig8}
\end{figure*}

The Confusion Matrix (CM) of all models is shown in Figure \ref{fig8}. It can be seen that the LSTM-GRU model has much misclasfiaciton, making it the worst performer compared to the other classifiers. The model \textit{HyExDNN-RNN} performed best with only one misclassification, indicating that this model should be used for ADHD detection and a better understanding of the decisions.

\subsubsection{Multi-Class Classification Results}
The results of the multi-class classification test, shown in Table \ref{table12}, provide useful information about the performance of each model. Among these models, the \textit{HyExDNN-RNN} model shows the highest performance in all metrics, with 94.20\% accuracy, precision, recall and F1-score, indicating a balanced and very effective model for the task. The DNN model also performs very well, with 93.0\% across all metrics, indicating strong predictive capabilities. EXGB is another powerful model, especially with an accuracy of 88.50\% and a corresponding F1 score, making it a reliable choice for this classification problem. The RF and DT models deliver decent performance but are outperformed by the more complex neural network-based models. The LSTM and LSTM-GRU models have lower accuracy and F1-score (83.0\% and 82.0\%, respectively), suggesting that these models are less effective compared to the others for this particular task. The LSTM-RNN model provides a balanced performance with an accuracy and recall of almost 86.0\% and a slightly lower F1-score of 85.0\%, showing moderate success in this classification task. It is also noted that the ROC-AUC of \textit{HyExDNN-RNN} is the highest for multi-class classification at 0.95.

\begin{table}[!ht]
\centering
\caption{Results of Multi-class Classification}
\label{table12}
\begin{tabular}{|c|c|c|c|c|c|}
\hline
\textbf{Models} & \textbf{Accuracy (\%)} & \textbf{Precision(\%)} & \textbf{Recall(\%)} & \textbf{F1-score(\%)}& ROC-AUC \\ \hline
RF & 87.23 & 82.45 & 87.22 & 84.50&0.88 \\ \hline
DT & 85.10 & 87.15 & 85.10 & 86.09 &0.86\\ \hline
EXGB & 88.50 & 87.90 & 88.50 & 88.50&0.89 \\ \hline
LSTM & 83.00 & 83.00 & 83.00 & 83.00&0.82 \\ \hline
DNN & 93.00 & 93.00 & 93.00 & 93.00&0.94 \\ \hline
LSTM-GRU & 82.00 & 82.00 & 82.00 & 82.00&0.82 \\ \hline
LSTM-RNN & 86.00 & 87.00 & 86.00 & 85.00&0.86 \\ \hline
\textit{HyExDNN-RNN} & 94.20 & 94.20 & 94.20 & 94.20&0.95\\ \hline
\end{tabular}
\end{table}

\begin{figure*}[!ht]
\centering
\subfloat[Accuracy of LSTM Model\label{fig2aLSTM}]{
\includegraphics[width=0.49\textwidth,height=4.9cm]{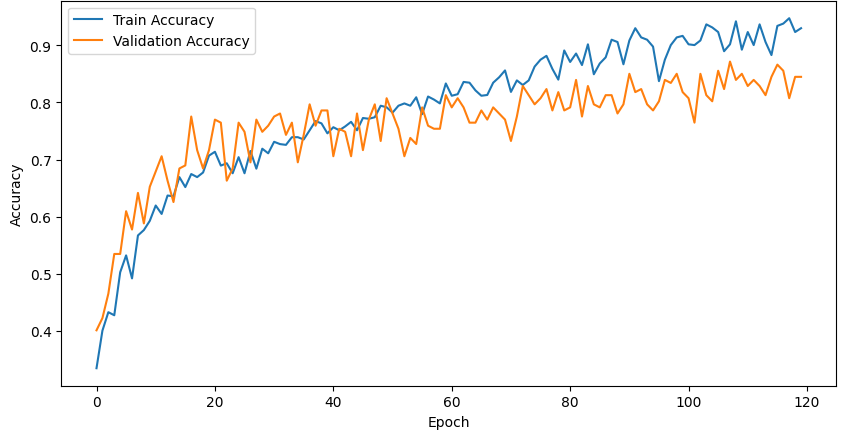} }
\subfloat[Loss of LSTM Model\label{fig2bLSTM}] {
\includegraphics[width=0.49\textwidth,height=4.9cm]{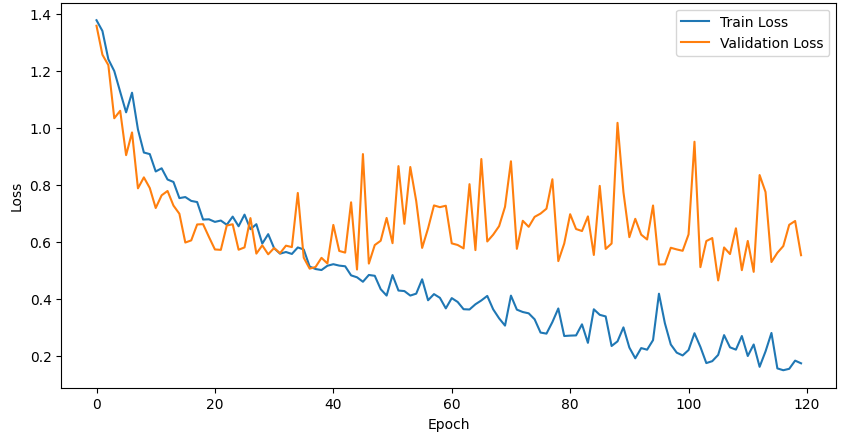}} 
\caption{Performance Visualization of LSTM Model}
\label{fig2LSTMaa}
\end{figure*}

Figure \ref{fig2LSTMaa} shows the performance of an LSTM model on a multi-class classification task. Figure \ref{fig2LSTMaa}(a) shows the accuracy of the model for both the training and test set, and Figure \ref{fig2LSTMaa}(b) shows the loss of the model for both the training and test sets. The LSTM model shows commendable performance by exhibiting high accuracy and low loss, as can be seen from the graphical representation. The accuracy graph shows a steady increase in training accuracy across epochs, exceeding the 90\% mark at epoch 100. The validation accuracy also shows a constant upward trend with little fluctuation, indicating that the model is learning effectively from the training set.
However, the frequent changes in the gap between the validation accuracy and the training accuracy compared to the training loss raise concerns that the training data has been over-fitted. This suggests that the model is learning the training data too well, which could lead to poor performance on the test data. This highlights the need for further investigation and possible improvements to the model. The training loss initially decreases and then gradually decreases, mirroring the fluctuations in validation loss. Although the latter initially decreases, it starts to increase around 0.5, indicating a possible overfitting to the training data. This emphasizes the importance of the loss graph for understanding the performance of the model.

\begin{figure*}[!ht]
\centering
\subfloat[Accuracy of DNN Model\label{fig5a}]{
\includegraphics[width=0.49\textwidth,height=4.9cm]{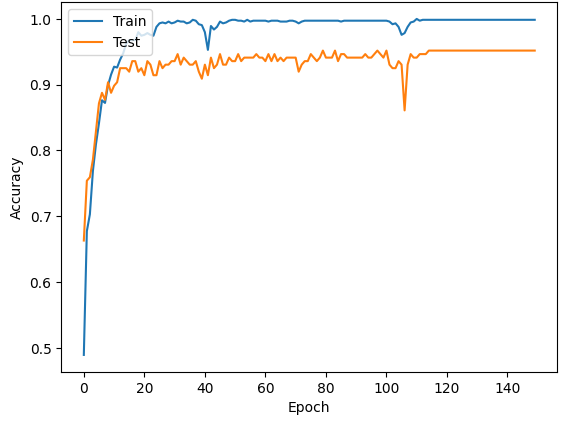}}
\subfloat[Loss of DNN Model\label{fig5b}]{
\includegraphics[width=0.49\textwidth,height=4.9cm]{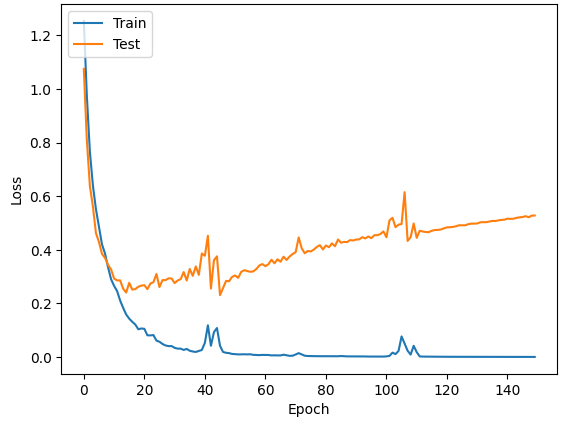}} 
\caption{Performance Visualization of DNN Model}
\label{fig5}
\end{figure*}

Figure \ref{fig5} shows the performance of a DNN model on a multi-class classification task. Figure \ref{fig5}(a) shows the accuracy of the model for both the training and test sets, and Figure \ref{fig5}(b) shows the loss of the model for both the training and test sets. The model achieves a high degree of accuracy on both the training and testing data, with training accuracy at 1.0 and test accuracy at 0.95. The loss of the model decreases rapidly in the first epochs, indicating that the model learns quickly. The loss eventually stabilizes, indicating that the model has converged to a good solution.

\begin{figure*}[!ht]
\centering
\subfloat[Accuracy of LSTM-GRU Model\label{fig4alstmgru}]{
\includegraphics[width=0.49\textwidth,height=4.9cm]{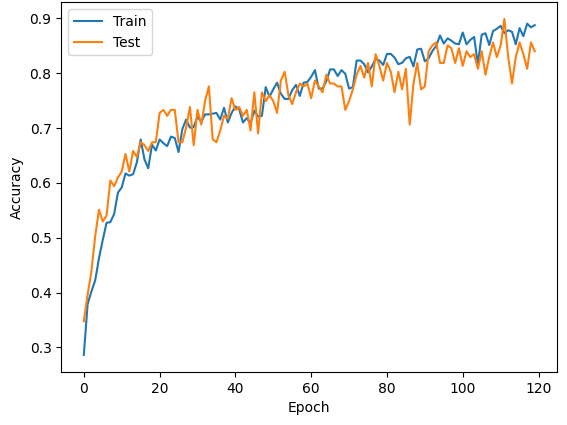} }
\subfloat[Loss of LSTM-GRU Model\label{fig4blstmgru}] {
\includegraphics[width=0.49\textwidth,height=4.9cm]{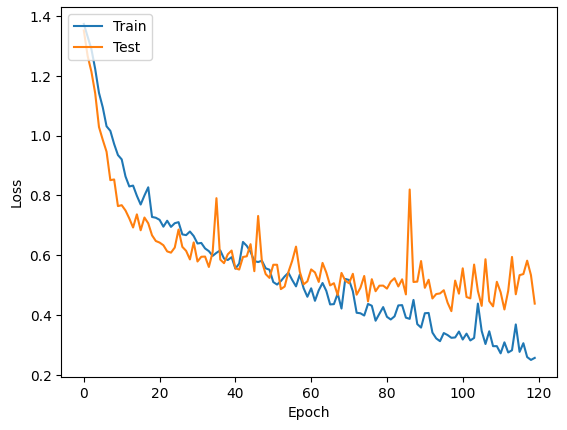}} 
\caption{Performance Visualization of LSTM-GRU Model}
\label{fig4lstmgru}
\end{figure*}

Figure \ref{fig4lstmgru} shows the performance of an LSTM-GRU model on a multi-class classification task. Figure \ref{fig4lstmgru}(a) shows the accuracy of the model for both the training set and the testing sets, and Figure \ref{fig4lstmgru}(b) shows the loss of the model for both the training set and the testing sets. The LSTM-GRU model works over 120 epochs. The accuracy of the model for both the training and test set increases with each epoch, reaching around 0.8. The loss graph shows that the loss for both the training and test sets decreases with each epoch, settling around 0.3.

\begin{figure*}[!ht]
\centering
\subfloat[Accuracy of LSTM-RNN Model\label{fig3ad}]{
\includegraphics[width=0.49\textwidth,height=4.9cm]{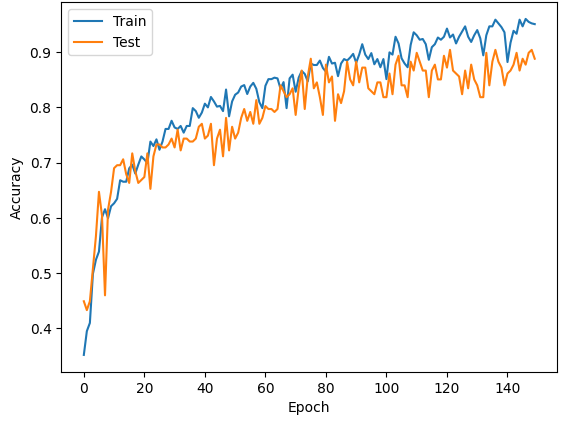} }
\subfloat[Loss of LSTM-RNN Model\label{fig3bd}] {
\includegraphics[width=0.49\textwidth,height=4.9cm]{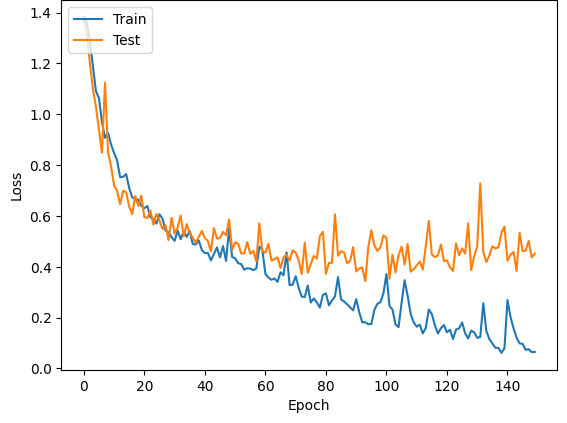}} 
\caption{Performance Visualization of LSTM-RNN Model}
\label{fig3lstmrnn}
\end{figure*}

Figure \ref{fig3lstmrnn} shows the performance of an LSTM-RNN model on a multi-class classification task. Figure \ref{fig3lstmrnn}(a) shows the accuracy of the model for both the training and test sets and Figure \ref{fig3lstmrnn}(b) shows the loss of the model for both the training and test sets. The LSTM-RNN training is characterized by a gradual increase in accuracy, which reaches 90\% for the training set, while the accuracy on the test set shows a less constant increase. This means that the model remembers the training data and fits it very well, but it might overfit and not generalize well on the test data. The loss curve indicates that the model 'memorizing' the data it was trained on (the loss decreases), while the testing loss is more volatile.

\begin{figure*}[!ht]
\centering
\subfloat[Training and Validation Accuracy of \textit{HyExDNN-RNN} model \label{m_dnnrnn}] {\includegraphics[trim={0 0 0 0.7cm},clip,width=0.5\textwidth]{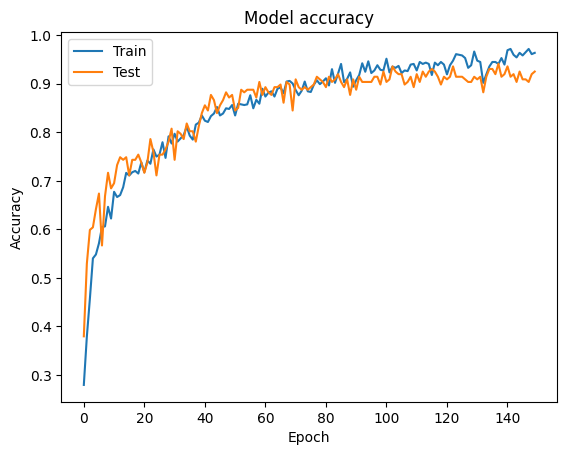}}
\subfloat[Training and Validation Loss of \textit{HyExDNN-RNN} model \label{m_dnnrnnloss}] {
\includegraphics[trim={0 0 0 0.7cm},clip,width=0.5\textwidth]{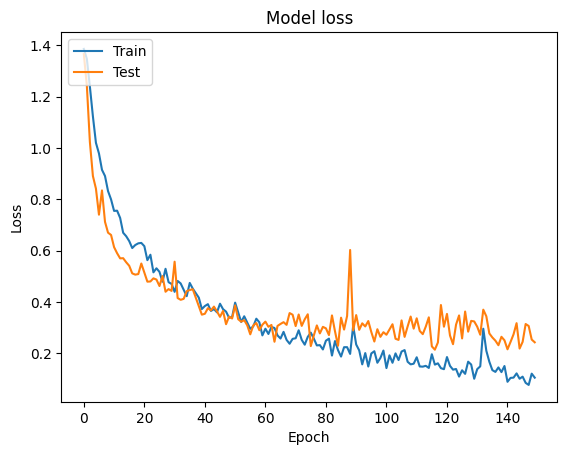}}\\
\subfloat[SHAP Interaction Values \label{m_shap}] {
\includegraphics[width=0.5\textwidth]{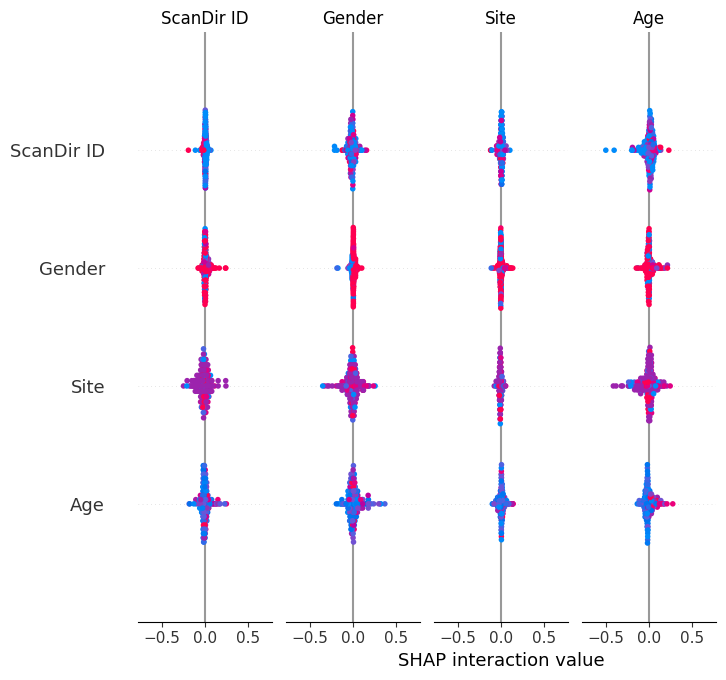}}
\subfloat[ADHD Features Extracted by PFI 
 \label{m_pfi}]{\includegraphics[trim={0 0 0 0.7cm},clip,width=0.5\textwidth]{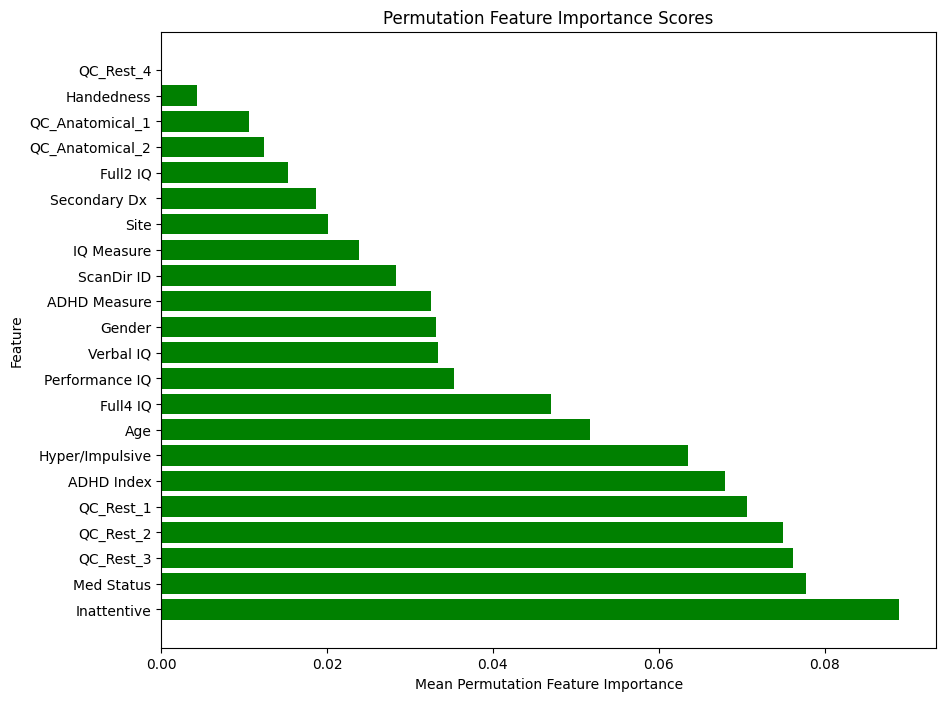}}
\caption{Performance Visualization of \textit{HyExDNN-RNN}  Multi-class Classification}
\label{multi}
\end{figure*}

Figure \ref{multi} shows the performance of a HyExDNN-RNN model on a multi-class classification task. Figure \ref{multi}(a) shows the accuracy of the model for both the training and testing sets, and Figure \ref{multi}(b) shows the loss of the model for both the training and test sets. The model consistently achieves an accuracy of approximately 94\% for the training data and approximately 94\% for the validation data. The fact that the training accuracy line is consistently above this and both lines show the same trend is a strong indication that the model is learning and generalizing correctly. The loss graph, which mirrors the accuracy graph, shows the improvement of the model over time. The training loss continues to decrease, while the steady decrease in training loss and the ability of the model to predict the correct output indicate the learning process of the model. The slightly higher loss in the validation data does not affect the overall improvement of the model. The model's ability to learn from the data is evident in the steadily decreasing loss and corresponding increase in accuracy, which emphasizes the model's strong performance. The SHAP interaction values for the model \textit{HyExDNN-RNN} are shown in Figure \ref{multi}(c). The SHAP interaction values are calculated by calculating the difference between the SHAP values of two features when they are considered together and separately. The SHAP interaction value for ScanDir ID is positive, indicating a positive interaction between ScanDir ID and Site. This suggests that a person with a ScanDir ID' has a greater influence on the model's prediction of whether they will fail alone than someone with a ScanDir ID. The SHAP interaction value for the Site is negative, indicating a negative interaction between ScanDir ID and the Site. This suggests that a person's Gender has less influence on the model's prediction of whether they would fail alone than their Gender with the Site. The SHAP interaction score for Gender is positive, indicating a positive interaction between ScanDir ID and Gender. This suggests that a person's Gender has a greater impact on the model's prediction of whether they will fail on their own than their ScanDir ID. It should be noted that these are only the expected properties of the \textit{HyExDNN-RNN} model. Depending on the dataset and the hyperparameters used the actual features that are most important for the prediction of the model may differ. Figure \ref{multi}(d) shows the PFI scores of the \textit{HyExDNN-RNN} model. The PFI scores are generated by randomly shuffling feature values and show how the predictions of the model vary. The features that have the greatest influence on the model's predictions are critical to the model's ability to produce a robust forecast. The features of the image are ordered from most to least important, with the most important feature at the top of the pyramid. QC\_Rest\_4, handedness, QC\_Anatomical\_1, QC\_Anatomical\_2, and Full 2 IQ Measure are the most important features. All of these characteristics are related to the subject's cognitive abilities. The QC\_Rest\_4 score assesses the subject's attention and concentration, the QC\_Anatomical\_2 score assesses the subject's brain anatomy, and the IQ Measure assesses the subject's intelligence. The other elements of the pyramid are also linked to the subject's cognitive abilities and physical and emotional well-being. The Handedness feature, for example, indicates whether the subject is left-handed or right-handed, while the QC Anatomical1 score measures the subject's brain connections, and the Full4 IQ score measures the subject's total cognitive function.

\begin{figure}[!ht]
\centering
\subfloat[DT model \label{m_conf_dt}] {
\includegraphics[width=0.33\columnwidth,height=4cm]{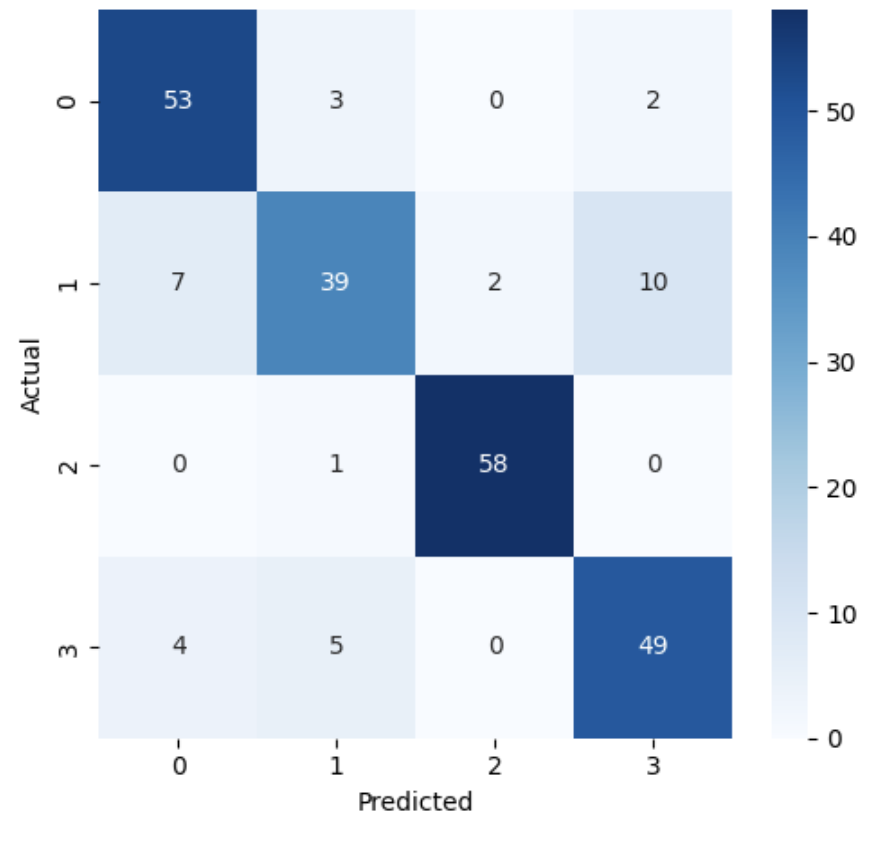}}
\subfloat[RF model \label{m_conf_rf}]{
\includegraphics[width=0.33\columnwidth,height=4cm]{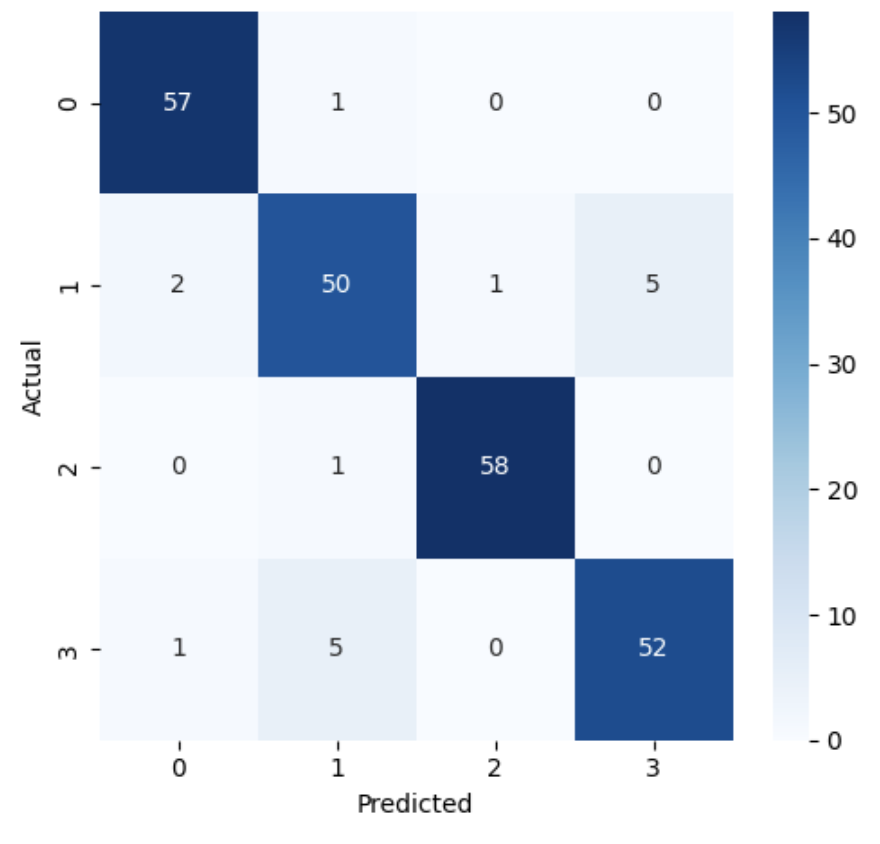}}
\subfloat[EXGB model \label{m_conf_xgb}] {
\includegraphics[width=0.33\columnwidth,height=4cm]{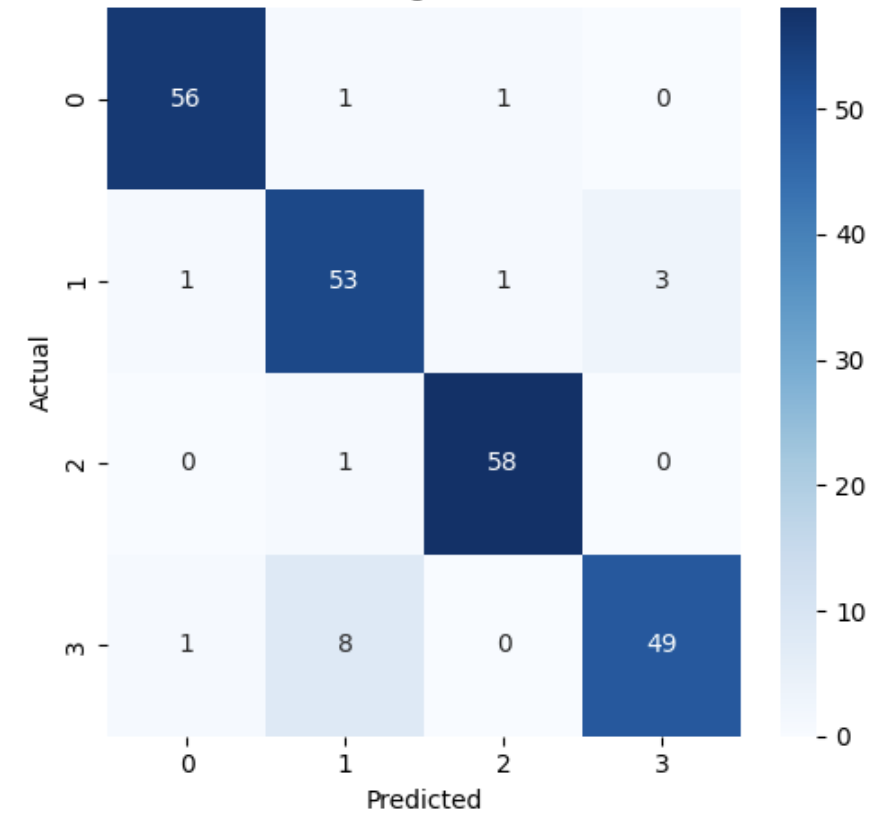}}\\
\subfloat[LSTM model \label{m_lstmcondasdaf}]{
\includegraphics[trim={0 0 0 0.7cm},clip,width=0.33\columnwidth,height=4cm]{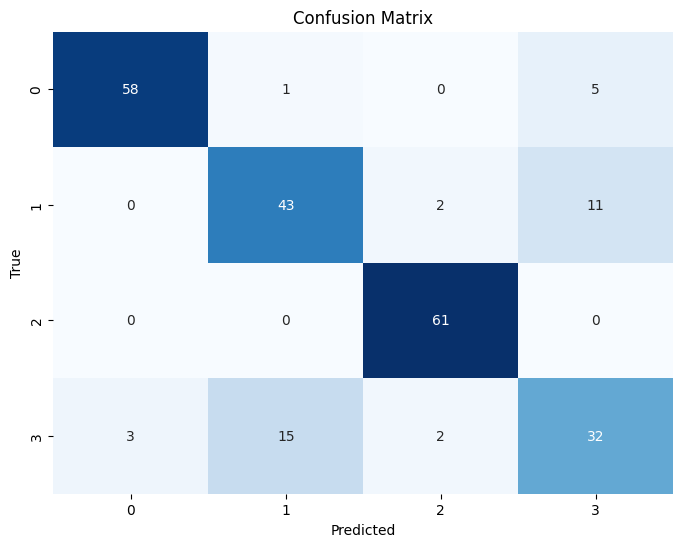}}
\subfloat[DNN model \label{m_lstmconf}]{
\includegraphics[trim={0 0 0 0.7cm},clip,width=0.33\columnwidth,height=4cm]{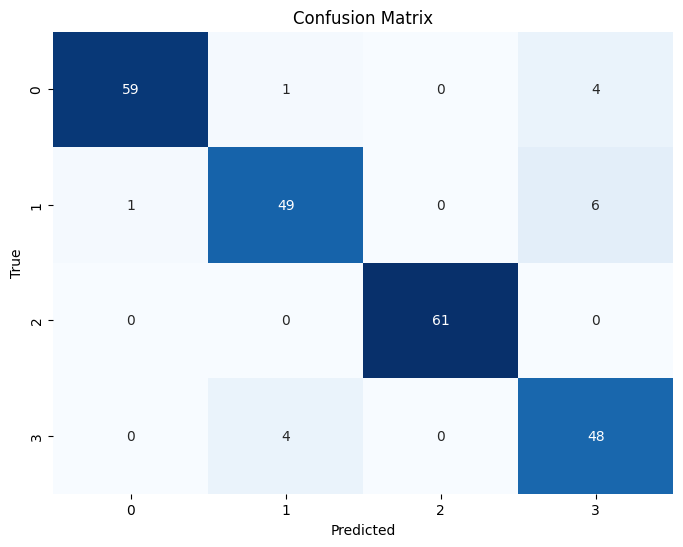}}
\subfloat[LSTM-GRU model \label{lstm}]{
\includegraphics[trim={0 0 0 0.7cm},clip,width=0.33\columnwidth,height=4cm]{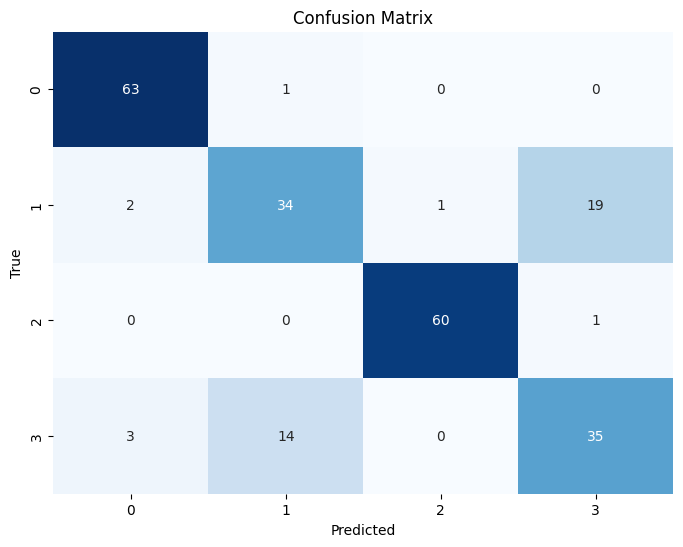}}\\
\subfloat[LSTM-RNN model \label{m_lstasdasdamconf}]{
\includegraphics[trim={0 0 0 0.7cm},clip,width=0.33\columnwidth,height=4cm]{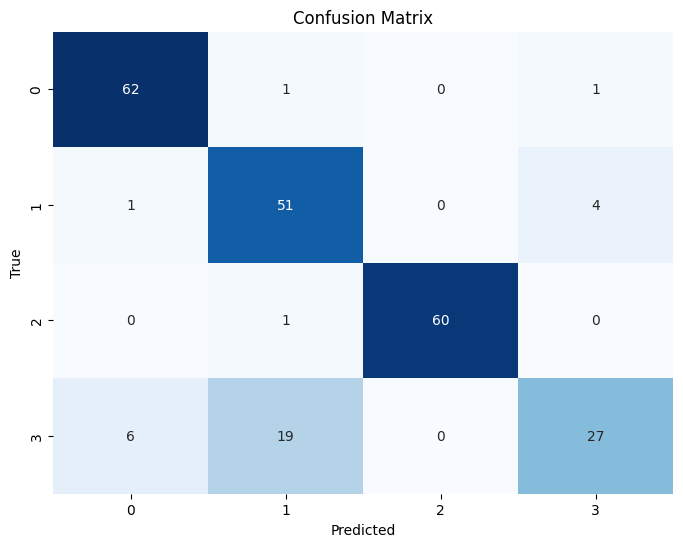}}
\subfloat[HyExDNN-RNN model \label{m_lcvsdvsdstmconf}]{
\includegraphics[trim={0 0 0 0.7cm},clip,width=0.33\columnwidth,height=4cm]{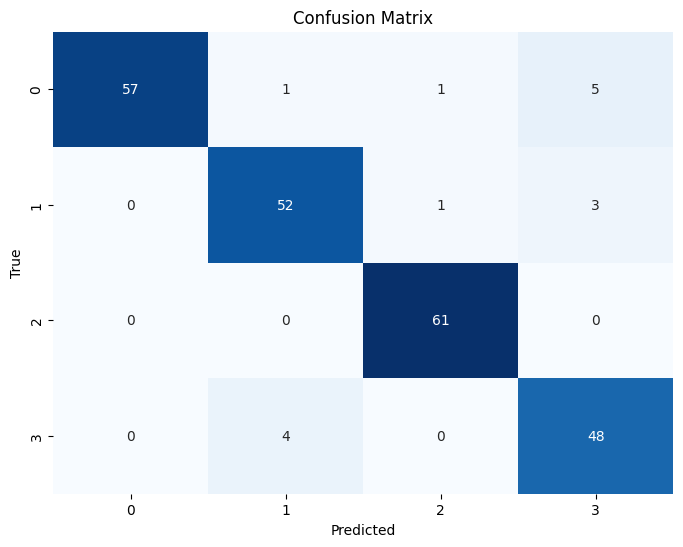}}
 \caption{Confusion Matrix for Multi-Class Classification}
\label{multi_confusion}
\end{figure}

Figure \ref{multi_confusion} shows the confusion matrices of all models. It can be seen that the LSTM and LSTM-GRU models have many misclassifications, making them the worst performers compared to the other classifiers. The reason for this could be that LSTMs perform well on sequential data, but this ADHD data is not sequential. In addition, the DNN and \textit{HyExDNN-RNN} models performed best with few misclassifications, suggesting that this model should be used for ADHD detection.

\subsection{Comparison with Existing Research}
In recent years, advances in machine learning techniques have played a critical role in changing healthcare practices, particularly in the diagnosis of complex neuropsychiatric disorders such as ADHD. Sharma et al.~\cite{sharma2023ipal} have contributed to this expanding field. This study is primarily concerned with the creation of IPAL, an intelligent framework for the automated diagnosis of ADHD using machine learning algorithms. The inspiration for this endeavour stems from the growing need for accurate and efficient diagnostic tools to help clinicians diagnose ADHD quickly and accurately. While Sharma's work provides a solid foundation, this paper aims to build upon it by combining XAI techniques with deep learning models to improve the accuracy and interpretability of ADHD categorization. We hope to shed light on the usefulness of XAI-driven deep learning models in solving the problems of ADHD diagnosis in this comparative analysis and highlight notable differences and improvements over the IPAL framework. The baseline model~\cite{sharma2023ipal} used various machine learning algorithms such as Logistic Regression (LR), Support Vector Machine (SVM), RF, K-Nearest Neighbours (KNN), AdBoost and Artificial Neural Network (ANN) for ADHD diagnosis. The results are shown in Table \ref{comparison}. LR has an accuracy of 86.13\%, SVM has a slightly higher accuracy of 90.09\%, RF has an accuracy of 84.15\%, KNN has a performance of 77.27\%, AdBoost has a performance of 69.30\%, and ANN has an accuracy of 85.14\%.
In comparison, the proposed \textit{HyExDNN-RNN} models outperformed all other models with 94.20\% accuracy. These results demonstrate the superiority of \textit{HyExDNN-RNN} in our framework with the best accuracy of all models, proving its utility in ADHD diagnosis. While SVM has shown competitive accuracy from previous studies, \textit{HyExDNN-RNN} has the added advantage of sequential data processing and improved interpretability, making it a promising alternative for ADHD diagnosis. This study advances the field by demonstrating the ability of deep learning models, in particular \textit{HyExDNN-RNN} augmented by XAI, to improve both the accuracy and interpretability of ADHD diagnosis compared to standard machine learning methods.

\begin{table}[!ht]
\caption{Comparison of Proposed and Existing Approach Results for Multi-class Classification}
\label{comparison}
\centering
\begin{tabular}{|cc|}
\hline
\multicolumn{1}{|c|}{\textbf{Models}} & \textbf{Accuracy} \\ \hline
\multicolumn{2}{|c|}{Existing Research Results~\cite{sharma2023ipal}} \\ \hline
\multicolumn{1}{|c|}{LR} & 86.13 \\ \hline
\multicolumn{1}{|c|}{SVM} & 90.09 \\ \hline
\multicolumn{1}{|c|}{RF} & 84.15 \\ \hline
\multicolumn{1}{|c|}{KNN} & 77.27 \\ \hline
\multicolumn{1}{|c|}{AdBoost} & 69.30 \\ \hline
\multicolumn{1}{|c|}{ANN} & 85.14 \\ \hline
\multicolumn{2}{|c|}{Proposed Framework Result} \\ \hline
\multicolumn{1}{|c|}{\textit{HyExDNN-RNN}} & 94.20 \\ \hline
\end{tabular}
\end{table}

\section{Discussion and Findings}\label{Discussion and Findings}

The motivation to involve psychologists in the process of explainable AI for ADHD detection stems from the need to bridge the gap between advanced AI models and practical, human-centred care. Psychologists have deep expertise in understanding the nuanced behaviours and symptoms associated with ADHD, which is critical for accurate diagnosis and personalized intervention. However, traditional AI models often operate as a "black box" and offer little transparency into how decisions are made. By involving psychologists in the development and application of explainable AI tools, we ensure that the technology not only aligns with clinical expertise but also enables psychologists to interpret and validate the results of the AI. This human-in-loop framework promotes confidence in decisions, improves diagnostic accuracy and supports more informed and holistic treatment planning, ultimately leading to better outcomes for individuals with ADHD. The proposed framework utilizes the XAI method to generate acceptable and intuitive explanations that can be used to examine the functioning of the model, the quality of the datasets, and the features used in the numerous ADHD datasets. We use the Pearson correlation, which measures the degree and direction of linear relationships between two variables. It assesses how strongly changes in one variable are reflected in changes in another, with values ranging from [0, 1]. The main objective of using Pearson correlation is to examine the degree of relationship between the features and to select the features with unique and non-redundant information. \textit{HyExDNN-RNN} models allow the integration of both spatial and temporal features, which makes them optimal for understanding complex patterns.
Moreover, \textit{HyExDNN-RNN} is better able to handle long-term dependencies, while DNN layers help to improve gradient flow through feature extraction, and it generalizes better across different tasks by capturing both static and dynamic aspects of the input data. The results of our binary classification task give us important insights into the performance of the different models. The EXGB, LSTM-RNN and \textit{HyExDNN-RNN} models achieved the highest performance metrics in accuracy, precision, recall and F1-score. These models show a superior ability to correctly identify both positive and negative classes, making them the most reliable options for this classification task. Also, for multi-class classification, \textit{HyExDNN-RNN} performed the best. This model is very effective and is particularly appreciated for its interpretability and ease of use.

\section{Conclusion} \label{conclusion}
In this paper, we propose an XAI-based framework integrated with machine and deep learning methods to diagnose ADHD individuals and help psychologists validate the results against their clinical knowledge, improve clinical judgment, enhance patient communication, identify areas for further investigation, and create improvement plans. Our main goal is to combine AI with human expertise to bridge the gap between advanced computational techniques and practical psychological applications. Using a standardized method that includes feature reduction, model selection and interpretation, we achieve accurate binary and multi-class classification. The results of our experiments confirm the efficiency of our recommended framework. \textit{HyExDNN-RNN} performed well in binary classification with an F1-score of 99\%. In addition, the XAI approaches SHAP and PFI provided important insights into the relevance of features and the decision logic of the model. These methods improved the interpretability of our deep learning models and made them transparent and usable for experts and non-experts alike. Overall, the proposed \textit{HyExDNN-RNN} improves the early and accurate diagnosis of ADHD by analyzing datasets to detect patterns that human observers might miss. \textit{HyExDNN-RNN} also excelled in multi-class classification with an accuracy of 94.20\%, \textit{HyExDNN-RNN} performed superiorly. \textit{HyExDNN-RNN} provides consistent and scalable assessments, improves treatment outcomes and reduces diagnostic bias. In the future, we will investigate the impact of different feature reduction strategies and XAI methodologies on the interpretability of the model. Furthermore, we intend to refine the pre-trained models for ADHD classification through transfer learning.

\section*{Conflict of interest}
The authors declare no conflict of interest.

\bibliography{ref}
\end{document}